\documentclass{article}
\usepackage[nonatbib, final]{nips_2017}
\usepackage[utf8]{inputenc} % allow utf-8 input
\usepackage[T1]{fontenc}    % use 8-bit T1 fonts
\usepackage{hyperref}       % hyperlinks
\usepackage{url}            % simple URL typesetting
\usepackage{booktabs}       % professional-quality tables
\usepackage{amsfonts}       % blackboard math symbols
\usepackage{nicefrac}       % compact symbols for 1/2, etc.
\usepackage{microtype}      % microtypography
\usepackage{graphicx}
\usepackage{amsmath}
\usepackage{bm}
\usepackage{algorithmicx}
\usepackage{algorithm}% http://ctan.org/pkg/algorithms
\usepackage{algpseudocode}% http://ctan.org/pkg/algorithmicx
\usepackage[table]{xcolor}

\title{MiAMix: Enhancing Image Classification through a Multi-stage Augmented Mixed Sample Data Augmentation Method}

\author{%
  Wen Liang \\
  Google Inc. \\
  Mountain View, CA 94043 \\
  \texttt{liangwen@google.com} \\
  \And
  Youzhi Liang \\
  Department of Computer Science \\
  Stanford University \\
  Stanford, CA 94305 \\
  \texttt{youzhil@stanford.edu} \\
  \AND
  Jianguo Jia \\
  Department of Computing \\
  Hong Kong Polytechnic University\\
  Hong Kong, China \\
  \texttt{jianguo1.jia@connect.polyu.hk}
}

\begin{document}

\maketitle

\begin{abstract}

Despite substantial progress in the field of deep learning, overfitting persists as a critical challenge, and data augmentation has emerged as a particularly promising approach due to its capacity to enhance model generalization in various computer vision tasks. While various strategies have been proposed, Mixed Sample Data Augmentation (MSDA) has shown great potential for enhancing model performance and generalization. We introduce a novel mixup method called MiAMix, which stands for Multi-stage Augmented Mixup. MiAMix integrates image augmentation into the mixup framework, utilizes multiple diversified mixing methods concurrently, and improves the mixing method by randomly selecting mixing mask augmentation methods. Recent methods utilize saliency information and the MiAMix is designed for computational efficiency as well, reducing additional overhead and offering easy integration into existing training pipelines. We comprehensively evaluate MiaMix using four image benchmarks and pitting it against current state-of-the-art mixed sample data augmentation techniques to demonstrate that MIAMix improves performance without heavy computational overhead.

\end{abstract}

\section{Introduction}

Deep learning has revolutionized a wide range of computer vision tasks like image classification, image segmentation, and object detection~\cite{resnet, unified}. However, despite these significant advancements, overfitting remains a challenge~\cite{out_of_distribution}. The data distribution shifts between the training set and test set may cause model degradation. This is also particularly exacerbated when working with limited labeled data or with corrupted data. Numerous mitigation strategies have been proposed, and among these, data augmentation has proven to be remarkably effective~\cite{randaugment, autoaug}. Data augmentation techniques increase the diversity of training data by applying various transformations to input images in the model training. The model can be trained with a wider slice of the underlying data distribution which improves model generalization and robustness to unseen inputs. Of particular interest among these techniques are mixup-based methods, which create synthetic training examples through the combination of pairs of training examples and their labels~\cite{mixup}. 

Subsequent to mixup, an array of innovative strategies were developed which go beyond the simple linear weighted blending of mixup, and instead apply more intricate ways to fuse image pairs. Notable among these are CutMix and FMix methods \cite{cutmix, fmix}. The CutMix technique \cite{cutmix} formulates a novel approach where parts of an image are cut and pasted onto another, thereby merging the images in a region-based manner. On the other hand, FMix \cite{fmix} applies a binary mask to the frequency spectrum of images for fusion, hence achieving an enhanced mixup process that can take on a wide range of mask shapes, rather than just square mask in CutMix. These methods have been successful in preserving local spatial information while introducing more extensive variations into the training data.

While mixup-based methods have shown promising results, there remains ample room for innovation and improvement. These mixup techniques utilize little to no prior knowledge, which simplifies their integration into training pipelines and incurs only a modest increase in training costs. To further enhance performance, some methodologies have leveraged intrinsic image features to boost the impact of mixup-based methods\cite{saliencymix}. Recently, following this approach, some methods employ the model-generated feature to guide the image mixing \cite{attentivemix}. Furthermore, some researchers have also incorporated image labels and model outputs in the training process as prior knowledge, introducing another dimension to improve these methods' performance\cite{automix}. The utilization of these methods often introduces a considerable increase in training costs to extract the prior knowledge and construct a mixing mask dynamically. This added complexity not only impacts the speed and efficiency of the training process but can also act as a barrier to deployment in resource-constrained environments. Despite their theoretical simplicity, in practice, these methods might pose integration challenges. The necessity to adjust the existing pipeline to accommodate these techniques could complicate the training process and hinder their adoption in a broader range of applications. Given this, we are driven to ponder an important question about the evolution of mixed sample data augmentation methods: How can we fully unleash the potential of MSDA while avoiding extra computational cost and facilitating seamless integration into existing training pipelines?

Considering the RandAugment\cite{randaugment} and other image augmentation policies, we are actually applying multiple layers of data augmentation to the input images and those works have shown that a multi-layered and diversified data augmentation strategy can significantly improve the generalization and performance of deep learning models. The work RandomMix\cite{randommix} starts ensembling the MSDA methods by randomly choosing one from a set of methods. However, by restricting to only one mixing mask can be applied, RandomMix imposes some unnecessary limitations. Firstly, the variety of mixing methods can be highly improved if multiple mask methods can be applied together. Secondly, the diversity of possible mixing shapes can be greater if we can further augment the mixing masks. Thirdly, we draw insights from AUGMIX, an innovative approach that apply different random sampled augmentation on the same input image and mix those augmented images. With the help of customized loss function design, it achieved substantial improvements in robustness. Inspired by this, we propose to remove a limitation in conventional MSDA methods and allow a sample to mix with itself with an assigned probability. It is essential to note that, during this mixing process, the input data must undergo two distinct random data augmentations.

In this paper, we propose the MiAMix: Multi-layered Augmented Mixup. MiAMIX alleviates the previously mentioned restrictions. Our contributions can be summarized as follows:

1. We firstly revisit the design of GMix\cite{gmix}, leading to an augmented form called AGMix. This novel form fully capitalizes the flexibility of Gaussian kernel to generate a more diversified mixing output.

2. A Novel sampling method of mixing ratio is designed for multiple mixing masks.

3. We define a new MSDA method with multiple stages: random sample paring, mixing methods and ratios sampling, generation and augmentation of mixing masks, and finally, the mixed sample output stage. We consolidate these stages into a comprehensive framework named MiAMix and establish a search space with multiple hyper-parameters.

To assess the performance of our proposed AGmix and MiAMix method, we conducted a series of rigorous evaluations across CIFAR-10/100, and Tiny-ImageNet\cite{tiny_imagenet} datasets. The outcomes of these experiments substantiate that MiAMix consistently outperforms the leading mixed sample data augmentation methods, establishing a new benchmark in this realm. In addition to measuring the generalization performance, we also evaluated the robustness of our model in the presence of natural noises. The experiments demonstrated that the application of RandomMix during training considerably enhances the model's robustness against such perturbations. Moreover, to scrutinize the effectiveness of our multi-stage design, we implemented an extensive ablation study using the ResNet18\cite{resnet} model on the Tiny-ImageNet dataset.

\section{Related Works}

Mixup-based data augmentation methods have played an important role in deep neural network training\cite{overview}. Mixup generates mixed samples via linear interpolation between two images and their labels\cite{mixup}. The mixed input $\tilde{x}$ and label $\tilde{y}$ are generated as:

\begin{equation}
\tilde{x} = \lambda x_i + (1 - \lambda) x_j,
\end{equation}
where $x_i$ , $x_j$ are raw input vectors.

\begin{equation}
\tilde{y} = \lambda y_i + (1 - \lambda) y_j,
\end{equation}
where $y_i$ , $y_j$ are one-hot label encodings.

$(x_i, y_i)$ and $(x_j, y_j )$ are two examples drawn at random from our training data, and $\lambda \in [0, 1]$. The $\lambda \sim \text{Beta}(\alpha, \alpha)$, for $\alpha \in (0, \infty)$. Following the development of Mixup, an assortment of techniques have been proposed that focus on merging two images as part of the augmentation process. Among these, CutMix\cite{cutmix} has emerged as a particularly compelling method.

In the CutMix approach, instead of creating a linear combination of two images as Mixup does, it generates a mixing mask with a square-shaped area, and the targeted area of the image are replaced by corresponding parts from a different image. This method is considered a cutting technique due to its method of fusing two images. The cutting and replacement idea has been also used in FMix\cite{fmix} and GridMix\cite{gridmix}.

The paper \cite{gmix} unified the design of different MSDA masks and proposed GMix. The Gaussian Mixup (GMix) generates mixed samples by combining two images using a Gaussian mixing mask. GMix first randomly selects a center point $c$ in the input image. It then generates a Gaussian mask centered at $c$, where the mask values follow:

\begin{equation}
mask_{gmix} = 1 - \exp\left(-\frac{|p-c|^2}{2\sigma^2}\right)
\end{equation}

where $\sigma$ is set based on the mixing ratio $\lambda$ and image size $N$ as

\begin{equation}
\sigma = \sqrt{\lambda} N
\end{equation}

This results in a smooth Gaussian mix of the two images, transitioning from one image to the other centered around the point $c$.

\section{Methods}

\subsection{GMix and Our AGMix}

\begin{figure}[ht]
  \includegraphics[width=120mm]{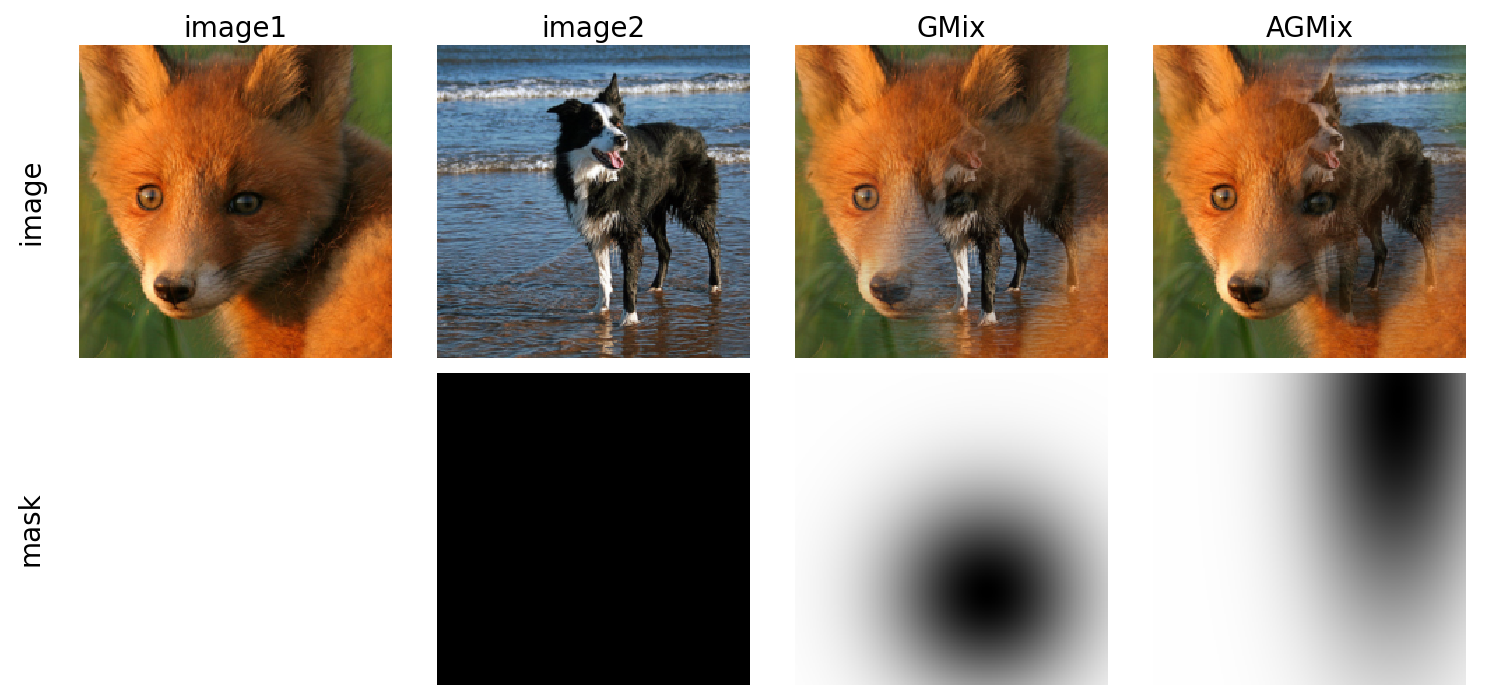}
  \centering
  \caption{Examples generated by GMix and AGMix. The first column shows the generated sample and the second row shows the corresponding mixing mask. We set $\lambda$ = 0.7 for both 2 methods.}
  \label{fig:agmix}
\end{figure}

To further enhance the mixing capabilities of our method, we extend the Gaussian kernel matrix used in GMix to a new kernel matrix with randomized covariance. The motivation behind this extension is to allow for more diversified output mixing shapes in the mix mask. Specifically, we replace the identity kernel matrix with a randomized kernel matrix as follows:

$$\large \Sigma = \begin{bmatrix}1 & q \\ q & 1\end{bmatrix} \qquad q \sim \mathcal{U}(-1,1)$$

Here, $\Sigma$ is the Gaussian kernel covariance matrix. We keep the value in the diagonal as 1, which means that we do not randomize the intensity of the mixing, which should be solely controlled by the mixing ratio coefficient $\lambda$. To preserve the assigned mixing ratio $\lambda$ and to constrain the shape of the mask region, we sample the parameter $q$ from a uniform distribution in a restricted range $(-1, 1)$. By randomizing the off-diagonal covariance $q$, we allow the mixing mask to have a broader range of shapes and mixing patterns. To add further variation to the mixing shape, we apply sinusoidal rotations to the mixing mask by defining a rotation matrix $R$ as follows:

\begin{equation}
R = \begin{pmatrix}
\cos \theta & -\sin \theta \\
\sin \theta & \cos \theta \
\end{pmatrix},
\end{equation}

where $\theta$ is a random rotation angle. We then rotate the mixing mask $M$ using the rotation matrix $R$ to obtain a rotated mixing mask $M_{rot}$ as follows:

\begin{equation}
M_{rot} = RMR^T.
\end{equation}

A comparative visualization between GMix and AGMix is depicted in Figure~\ref{fig:agmix}. This comparison underlines the successful augmentation of the original GMix approach by AGMix, introducing a wealth of varied shapes and distortions. This innovation also inspires us to apply similar rotational and shear augmentations to other applicable mixing masks. In the forthcoming experiment results section, a series of experiments provides an in-depth comparison of AGMix and GMix, further underscoring the enhancements and improvements brought by the method.

\subsection{MiAMix}

\begin{figure}[ht]
  \includegraphics[width=120mm]{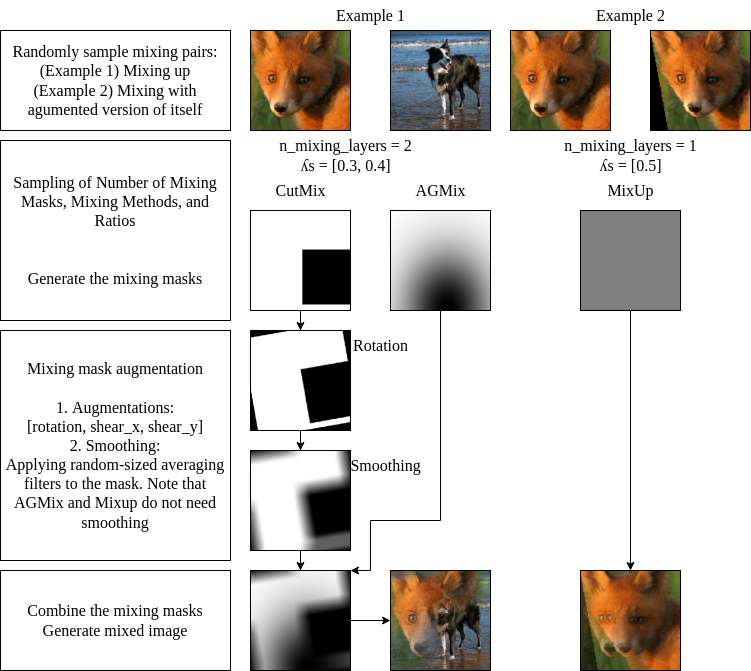}
  \centering
  \caption{An illustrative example of the MiAMix process, involving: 1) Random sample pairing; 2) Sampling the number, methods, and ratios of mixing masks; 3) Augmentation of mixing masks; 4) Generation of the final mixed output.}
  \label{fig:algo}
\end{figure}

We introduce the MiAMix method and its detailed designs in this section. The framework is constructed by 4 distinct stages: random sample paring, sampling of mixing methods and ratios, the generation and augmentation of mixing masks, and finally, the mixed sample output stage. Each stage will be discussed in the ensuing subsections. These stages are presented step-by-step in Algorithm~\ref{alg:miamix}, the parameters are listed in Table~\ref{table:MiAMix Parameters}, and a practical illustration of the processes within each stage can be found in Figure~\ref{fig:algo}.

To understand the effects of the various design choices of this proposed algorithm in this section, we conduct a series of ablation studies in the following experiment result section. We also compare our method with previous MSDA methods to justify that the MiAMix works as a balance between performance and computational overhead.

\begin{algorithm}[h]
\caption{Multi-stage Augmented Mixup (MiAMix)}
\label{alg:miamix}
\begin{algorithmic}[1]
\State \textbf{Inputs:} Data samples $x_1, x_2, ..., x_n$, corresponding labels $y_1, y_2, ..., y_n$, 
\State \textbf{Parameters:} mixing parameter $\alpha$, maximum number of mixing layers $k_{max}$, mixing method candidates $M$ and corresponding sampling weights $W$, more parameters are listed in the Table~\ref{table:MiAMix Parameters}
\State \textbf{Outputs:} Mixed samples $\tilde{x}_1, \tilde{x}_2, ..., \tilde{x}_n$, mixed labels $\tilde{y}_1, \tilde{y}_2, ..., \tilde{y}_n$
\State \For{$i=1$ to $n$}
\State Sample a mixing data point $(x_t, y_t)$ either by sampling from the entire pool of data samples or alternatively, selecting itself as the mixing data point with a ratio $p_{self}$.
\State Sample number of mixing layers $k$ from 1 to $k_{max}$
\State Sample $\lambda_1, \lambda_2, \ldots, \lambda_{k}$ from a Dirichlet distribution $\text{Dir}(\boldsymbol{\alpha})$, where the parameter vector $\boldsymbol{\alpha} = (\alpha_1, \ldots, \alpha_k, \alpha_{k+1})$, such that $\alpha_1 = \alpha_k = \alpha$ and $\alpha_{k+1} = k \cdot \alpha$.
\State Sample $k$ mixing methods $m_1$, $m_2$, ..., $m_k$ from $M$ with weighted distribution over $W$
\State Generate all $mask_{j}$ from $m_{j}(\lambda_{j})$
\State Apply mask augmentation to masks
\State Merge all the $k$ masks to $mask_{merged}$, Get the $\lambda_{merged}$ from the $mask_{merged}$
\State Apply $m_{merged}$ to the sampled input pair $\tilde{x}_i = mask_{merged} \otimes x_i + (1 - mask_{merged}) \otimes x_t$
\State Apply $\lambda_{merged}$ to sampled label pair $\tilde{y}_i = \lambda y_i + (1 - \lambda) y_j$
\State Append mixed $\tilde{x}_i$ and $\tilde{y}_i$ to output list
\EndFor
\State \textbf{return} Mixed samples $\tilde{x}_1, \tilde{x}_2, ..., \tilde{x}_n$, mixed labels $\tilde{y}_1, \tilde{y}_2, ..., \tilde{y}_n$
\end{algorithmic}
\end{algorithm}

\begin{table}[h]
\centering
\caption{MiAMix Parameters}
\begin{tabular}{|c|c|l|}
\hline
\textbf{Notation} & \textbf{Value} & \textbf{Description} \\
\hline
$\alpha$ & 1 & MSDA mix ratio sampling parameter \\
$k_{max}$ & 2 & Maximum number of mixing layers \\
$M$ & {[MixUp, CutMix, FMix, GridMix, AGMix]} & Mixing method candidates \\
$W$ & {[2, 1, 1, 1, 1]} & Mixing method sampling weights \\
$p_{self}$ & 0.10 & Self-mixing ratio \\
$p_{aug}$ & 0.25 & Mixing mask augmentation ratio \\
$p_{smooth}$ & 0.5 & Mixing mask smoothing ratio \\
\hline
\end{tabular}
\label{table:MiAMix Parameters}
\end{table}

\subsubsection{Random Sample Paring}

The conventional method of mix pair sampling is direct shuffling the sample indices to establish mixing pairs. There are two primary differences that arise in our approach. The first difference is that, in our image augmentation module, we prepare two sets of random augmentation results for mixing. If all images within a batch undergo the exact same augmentation, the ultimate mix's diversity remains constrained. This observation, inspired by our examination of the open-source project OpenMixup\cite{openmixup}, revealed a crucial oversight in prior work. In MiAMix, we addressed this issue and yielded measurable improvement. The second, and arguably more critical distinction, is the introduction of a new probability parameter, denoted as $p_{self}$, which enables images to mix with themselves and generate "corrupted" outputs. This strategy draws from the notable enhancement in robustness exhibited by AUGMIX\cite{augmix}. Integrating the scenario of an image mixing with itself can significantly benefit the model, and we delve into an experimental section of this paper.

\subsubsection{Sampling Number of Mixing Masks, Mixing Methods, and Ratios}

Previous studies such as RandAug and AutoAug have shown that ensemble usage and multi-layer stacking in image data augmentation are essential for improving a computer vision model and mitigating overfitting\cite{randaugment}. However, the utilization of ensembles and stacking in mixup-based methods has been underappreciated. Therefore, to enhance input data diversity with mixing, we introduce two strategies. Firstly, we perform random sampling over different methods. For each generation of a mask, a method is sampled from a mixing methods set $M$, with a corresponding set of sampling weights $W$. The $M$ contains not only our proposed method AGMix above but also MixUp, CutMix, GridMix and FMix. These mixup techniques blend two images with varying masks, and the main difference between those methods is how it generates these randomized mixing masks. As such, an MSDA can be conceptualized as a standardized mask generator, denoted by $m$. This generator takes as input a designated mixing ratio, $\lambda$, and outputs a mixing mask. This mask shares the same dimensions as the original image, with pixel values ranging from 0 to 1. And the final image can be directly procured using the formula: 

\begin{equation}
\tilde{x} = mask \otimes x_1 + (1 - mask) \otimes x_2
\end{equation}

In this context, $\otimes$ denotes element-wise multiplication, the mask is the generated mixing mask, and $x_1$ and $x_2$ represent the 2 original images.

Secondly, We pioneer the integration of multi-layer stacking in mixup-based methods. Therefore, we need to sample another parameter to set the mixing ratio for each mask generation step. For this, the mixup's methodology here is:

\begin{equation}
\lambda \sim \text{Beta}(\alpha, \alpha), \text{for} \alpha \in (0, \infty)
\end{equation}

While the Beta distribution's original design caters to bivariate instances, the Dirichlet distribution presents a multivariate generalization. It's a multivariate probability distribution parameterized by a positive reals vector $\boldsymbol{\alpha}$, essentially generalizing the Beta distribution. Our sampling approach is:

\begin{equation}
\lambda_1, \lambda_2, \ldots, \lambda_{k} \sim \text{Dir}(\boldsymbol{\alpha}), \quad \text{for $k$ masks} \nonumber
\end{equation}
\begin{equation}
\text{where } \boldsymbol{\alpha} = (\alpha_1, \ldots, \alpha_k, \alpha_{k+1})\text{, and } \alpha_1 = \alpha_k = \alpha, \alpha_{k+1} = k \times \alpha
\end{equation}

We maintain $\alpha$ as the sole sampling parameter for simplicity. With the Dirichlet distribution's multidimensional property, the mixing ratios derived from sampling are employed for multiple mask generators. In other words, our MiAMix approach employs the parameter $\lambda_i$ to determine the mixing ratio for each mask $mask_i$. This parameter selection method plays a pivotal role in defining the multi-layered mixing process.

\subsubsection{Mixing Mask Augmentation}

Upon generation of the masks, we further execute augmentation procedures on these masks. To preserve the mixing ratio inherent to the generated masks, the selected augmentation processes should not bring substantial change to the mixing ratio of the mask, so we mainly focus on some morphological mask augmentations. Three primary methods are utilized: shear, rotation, and smoothing. The smoothing applies an average filter with varying window sizes to subtly smooth the mixing edge. It should be explicitly noted that these augmentations are particularly applicable to CutMix, FMix, and GridMix methodologies. In contrast, Mixup and AGMix neither require nor undertake the aforementioned augmentations.

\subsubsection{Mixing Output}

During the mask generation step, we may have multiple mixing masks. The MiAMix employs the masks to merge two images and obtains the mixed weights for labels by point-wise multiplication. 

\begin{equation}
\text{mask}_{product} = \prod_{i=1}^{n} \text{mask}_i
\end{equation}

The $n$ denotes the number of masks, and the multiplication operation is conducted in a pointwise manner. Another approach we also tried is by summing the weighted mask:

\begin{equation}
\text{mask}_{sum} = \text{clip}\left(\sum_{i=1}^{n} \text{mask}_i, 0, 1\right),
\end{equation}

$\text{clip}$ serves to confine the mixing ratio at each pixel within the [0,1] interval. It is crucial to note that the cumulative mask weights could potentially exceed 1 at specific pixels. As a consequence, we enforce a clipping operation subsequent to the summation of masks if we sum them up.

In the output stage, our approach is different from the conventional mixup method. We sum the weights of the merged mask, $mask_{merged}$, to determine the final $\lambda_{merged}$, which defines the weights of the labels. 

\begin{equation}
\lambda_{merged} = \frac{1}{H \times W} \sum_{j=1}^{H} \sum_{k=1}^{W} \text{mask}_{merged_{jk}}
\end{equation}

In this equation, $H$ and $W$ denote the height and width of the mask, respectively, $j$ and $k$ are the indices of the pixels within each mask. Therefore, $\lambda_{merged}$ represents the overall mixing intensity by averaging the mixing ratios over all the pixels in $mask_{merged}$. The rationale behind this is that, if multiple masks have significant overlap between them, the final mixing ratio will deviate from the initially set $\lambda_{sum} = \Sigma \lambda_i$, regardless of whether the masks are merged via multiplication or summation. We will compare these two ways of merging the mixing mask and two ways of acquiring the weights $\lambda$ for labels in the upcoming experimental results section.
\section{Results}

In order to examine the benefits of MiAMix, we conduct experiments on fundamental tasks in image classification. Specifically, we chose the CIFAR-10, CIFAR-100, and Tiny-ImageNet datasets for comparison with prior work. We replicate the corresponding methods on all those datasets to demonstrate the relative improvement of employing this method over previous mixup-based methods.

\subsection{Tiny-ImageNet, CIFAR-10 and CIFAR-100 Classification}

For our image classification experiments, we utilize the Tiny-ImageNet\cite{tiny_imagenet} dataset, which consists of 200 classes with 500 training images and 50 testing images per class. Each image in this dataset has been downscaled to a resolution of $64 \times 64$ pixels. We also evaluate our methods (AGMix and MiAMix) against those mixing methods on CIFAR-10 and CIFAR-100 datasets. The CIFAR-10 dataset consists of 60,000 32x32 pixel images distributed across 10 distinct classes, and the CIFAR-100 dataset, mirroring the structure of the CIFAR-10 but encompasses 100 distinct classes, each holding 600 images. Both datasets include 50,000 training images and 10,000 for testing.

Training is performed using ResNet-18 and ResNeXt-50 network architecture over the course of 400 epochs, with a batch size of 128. Our optimization strategy employs Stochastic Gradient Descent (SGD) with a momentum of 0.9 and weight decay set to $5 \times 10^{-4}$. The initial learning rate is set to 0.1 and decays according to a cosine annealing schedule.

In our investigation of various mixup methods, we select a set of methods $M = [Mixup, CutMix, Fmix, GridMix, AGMix]$. Each of these methods was given a weight, represented as a vector $\mathbf{W} = [2, 1, 1, 1, 1]$. The mixing parameter, $\alpha$, was set to 1 throughout the experiments.

As shown in Table~\ref{table:image classification accuracy}, we compare the performance and training cost of several MSDA methods. The training cost is measured as the ratio of the training time of the method to the training time of the vanilla training. From the results, it is clear that our proposed method, MiAMix, shows a state-of-the-art performance among those low-cost MSDA methods. The test results even surpass the AutoMix which embeds the mixing mask generation into the training pipeline to take more advantage of injecting dynamic prior knowledge into the sample mixing. Notably, the MiAMix method only incurs an 11\% increase in training cost over the vanilla model, making it a cost-effective solution for data augmentation. In contrast, the AutoMix takes approximately 70\% more training costs.
\begin{table}[ht]
\centering
\scriptsize
\caption{Comparison of various MSDA on CIFAR-10 and CIFAR 100 using ResNet-18 and ResNeXt-50 backbones, on Tiny-ImageNet using a ResNet-18 backbone. Note that AutoMix needs additional computations for learning and processing extra prior knowledge. $Training\text{ }Cost = \frac{Training\text{ }time}{Vanilla\text{ }model\text{ }training\text{ }time}$}
\begin{tabular}{lcccccc}
\hline
& \multicolumn{2}{c}{\textbf{CIFAR10}} & \multicolumn{2}{c}{\textbf{CIFAR100}} & \textbf{Tiny-ImageNet} & \textbf{Training Cost} \\
\textbf{Methods} & \textbf{ResNet-18(\%)} & \textbf{ResNeXt-50(\%)} & \textbf{ResNet-18(\%)} & \textbf{ResNeXt-50(\%)} & \textbf{ResNet-18(\%)} \\
\hline
Vanilla & 95.07 & 95.81 & 77.73 & 80.24 & 61.68 & 1.00 \\
Mixup\cite{mixup} & 96.35 & 97.19 & 79.34 & 81.55 & 63.86 & 1.00 \\
CutMix\cite{cutmix} & 95.93 & 96.63 & 79.58 & 78.52 & 65.53 & 1.00 \\
FMix\cite{fmix} & 96.53 & 96.76 & 79.91 & 78.99 & 63.47 & 1.07 \\
GridMix\cite{gridmix} & 96.33 & 97.30 & 78.60 & 79.80 & 65.14 & 1.03 \\
GMix\cite{gmix} & 96.02 & 96.25 & 78.97 & 78.90 & 64.41 & 1.00 \\
SaliencyMix\cite{saliencymix} & 96.36 & 96.89 & 79.64 & 79.72 &  64.60 & 1.01 \\
\textcolor{gray}{AutoMix\cite{automix}} & \textcolor{gray}{97.08} & \textcolor{gray}{97.42} & \textcolor{gray}{81.78} & \textcolor{gray}{83.32} & \textcolor{gray}{67.33} & \textcolor{gray}{1.87} \\
\rowcolor{gray!30} AGMix & 96.15 & 96.37 & 79.36 & 81.04 & 65.68 & 1.03 \\
\rowcolor{gray!30} MiAMix & \textbf{96.92} & \textbf{97.52} & \textbf{81.43} & \textbf{83.50} & \textbf{67.95} & 1.11 \\
\hline
\end{tabular}
\label{table:image classification accuracy}
\end{table}

\subsection{Robustness}

To assess robustness, we set up an evaluation on the CIFAR-100-C dataset, explicitly designed for corruption robustness testing and providing 19 distinct corruptions such as noise, blur, and digital corruption. Our model architecture and parameter settings used for this evaluation are consistent with those applied to the original CIFAR-100 dataset in our above experiments. According to Table~\ref{table:corrupted classification accuracy}, our proposed MiAMix method demonstrated exemplary performance, achieving the highest accuracy. This provides compelling evidence that our multi-stage and diversified mixing approach contributes significantly to the improvement of model robustness.

\begin{table}[ht]
\centering
\caption{Top-1 accuracy on CIFAR-100 and corrupted CIFAR-100-C based on ResNeXt-50}
\begin{tabular}{lcc}
\hline
\textbf{Methods} & \textbf{Clean Acc(\%)} & \textbf{Corrupted Acc(\%)} \\
\hline
Vanilla & 80.24 & 51.71 \\
Mixup\cite{mixup} & 81.55 & 58.10 \\
CutMix\cite{cutmix} & 78.52 & 49.32 \\
\textcolor{gray}{AutoMix\cite{automix}} & \textcolor{gray}{83.32} & \textcolor{gray}{58.36} \\
\rowcolor{gray!30} MiAMix & 83.50 & \textbf{58.99} \\
\hline
\end{tabular}
\label{table:corrupted classification accuracy}
\end{table}

\subsection{Ablation Study}

The MiAMix method involves multiple stages of randomization and augmentation which introduce many parameters in the process. It is essential to clearly articulate whether each stage is necessary and how much it contributes to the final result. Furthermore, understanding the influence of each major parameter on the outcome is also crucial. To further demonstrate the effectiveness of our method, we conducted several ablation experiments on the CIFAR-10, CIFAR-100-C and Tiny-ImageNet datasets.

\subsubsection{GMix, AGMix, and Mixing Mask Augmentation}

A particular comparison of interest is between the GMix and our augmented version, AGMix in Table~\ref{table:image classification accuracy} and Table~\ref{table:corrupted classification accuracy}. The primary difference between these two methods lies in the inclusion of additional randomization in the Gaussian Kernel. The experiment results reveal that this simple yet effective augmentation strategy indeed brings about a significant improvement in the performance of the mixup method across all three datasets and one corrupted dataset, despite maintaining almost the same training cost as GMix. As the results in Table~\ref{table:ablation_study_augmentation} illustrate, the introduction of various forms of augmentation progressively improves model performance. These experiment results underscore the importance and effectiveness of augmenting mixing masks during the training process, furthermore, validate the approach taken in the design of our MiAMix method. 

\begin{table}[ht]
\centering
\caption{Ablation study on mixing mask augmentation with ResNet-18 on Tiny-ImageNet. The percentage after "Smoothing" and "rotation and shear" refers to the ratio of masks applied with the respective type of augmentation during training.}
\begin{tabular}{lcc}
\hline
\textbf{Augmentations} & \textbf{Top-1(\%)} & \textbf{Top-5(\%)} \\
\hline
No augmentation & 66.87 & 86.66 \\
+Smoothing 50\% & 67.29 & 86.82 \\
+rotation and shear 25\% & \textbf{67.95} & \textbf{87.26} \\
\hline
\end{tabular}
\label{table:ablation_study_augmentation}
\end{table}

\subsubsection{The Effectiveness of Multiple Mixing Layers}

\begin{table}[ht]
\centering
\caption{Ablation study on multiple mixing layers with ResNet-18 on Tiny-ImageNet. The brackets indicate that the number of turns is randomly selected from the enclosed numbers with equal probability during each training step.}
\begin{tabular}{lcc}
\hline
\textbf{Number of Turns} & \textbf{Top-1 (\%)} & \textbf{Top-5 (\%)} \\ 
\hline
1 & 66.16 & 86.49 \\
2 & \textbf{67.10} & 86.45 \\
3 & \textbf{67.10} & 86.42 \\
4 & 67.01 & 86.38 \\
$[1, 2]^*$ & \textbf{67.95} & \textbf{87.25} \\
$[1, 2, 3]^*$ & 67.86 & 87.16 \\
\hline
\end{tabular}
\label{table:ablation_study_num_layers}
\end{table}

The data presented in Table~\ref{table:ablation_study_num_layers} demonstrates the substantial impact of multiple mixing layers on the model's performance. As the table shows, a discernible improvement in Top-1 accuracy is observed when more layers of masks are added, emphasizing the effectiveness of this approach in enhancing the diversity and complexity of the training data. Most notably, the mod el's performance is further amplified when the number of layers is not constant but rather sampled randomly from a set of values, as indicated by the bracketed entries in the table. This observation suggests that introducing variability in the number of mixing layers could potentially be an effective approach for extracting more comprehensive and robust features from the data.

\subsubsection{The Effectiveness of MSDA Ensemble}

\begin{table}[h]
\centering
\caption{Effectiveness experiment of MSDA ensemble, tested on CIFAR-10 dataset. Each weight corresponds to a different MSDA candidate, and a weight of zero signifies the removal of the corresponding method from the ensemble.}
\begin{tabular}{cc}
\hline
\textbf{Weights {[}MixUp, CutMix, FMix, GridMix, AGmix{]}} & \textbf{Top-1 Accuracy (\%)} \\
\hline
{[}1, 1, 1, 1, 1{]} & \textbf{96.86} \\
{[}0, 1, 1, 1, 1{]} & 96.42 \textcolor{red}{-0.44}\\
{[}1, 0, 1, 1, 1{]} & 96.74 \textcolor{red}{-0.12}\\
{[}1, 1, 0, 1, 1{]} & 96.65 \textcolor{red}{-0.21}\\
{[}1, 1, 1, 0, 1{]} & 96.67 \textcolor{red}{-0.19}\\
{[}1, 1, 1, 1, 0{]} & 96.53 \textcolor{red}{-0.33}\\
\hline
\end{tabular}
\label{tab:msda_ensemble}
\end{table}

In the study, the ensemble's efficacy was tested by systematically removing individual mixup-based data augmentation methods from the ensemble and observing the impact on Top-1 accuracy. The results, as shown in Table~\ref{tab:msda_ensemble}, clearly exhibit the vital contributions each method provides to the overall performance. Eliminating any single method from the ensemble led to a decrease in accuracy, underscoring the value of the diverse mixup-based data augmentation techniques employed. This demonstrates the strength of our MiAMix approach in harnessing the collective contributions of these diverse techniques, optimizing their integration, and achieving superior performance results.

\subsubsection{Comparison Between Mask Merging Methods and Mixing Ratio Merging Methods}

\begin{table}[ht]
\centering
\caption{Comparison between different ways of merging multiple mixing masks and merging mixing ratios on Tiny-ImageNet with a ResNet-18 model. "sum" and "mul" respectively refer to merging masks through sum and multiplication. "merged" and "orig" denote the methods of acquiring $\lambda$ – either averaging the final merged mask or reusing the original $\lambda$.}
\begin{tabular}{cccc}
\hline
\textbf{Mask merge method} & \textbf{lambda merge method} & \textbf{Top-1(\%)} & \textbf{Top-5(\%)} \\
\hline
 mul & merged & \textbf{67.95} & \textbf{87.26} \\
 sum & merged & 66.58 \textcolor{red}{-0.37} & 86.60 \\
 mul & orig & 66.42 \textcolor{red}{-0.53} & 85.89 \\
\hline
\end{tabular}
\label{table:merge_methods}
\end{table}

As shown in Table~\ref{table:merge_methods}, the combination of multiplication for mask merging and the "out" method for $\lambda$ merging yields the highest accuracy for both Top-1 (67.95\%) and Top-5 (87.26\%). On the other hand, when using the sum operation for mask merging or reusing the original $\lambda$ (the "orig" method), the performance degrades. This suggests that reusing the original $\lambda$ might not provide a sufficiently adaptive mixing ratio for the model's learning process. Moreover, compared with the multiplication operation, the lower flexibility of the sum operation does impede the performance. These results reaffirm the superiority of the (mul, out) method in our multi-stage data augmentation framework.

\subsubsection{Effectiveness of Mixing with an Augmented Version of the Image Itself}

\begin{table}[ht]
\centering
\caption{Impact of self-mixing ratio on CIFAR-100 and CIFAR-100-C with ResNeXt-50. "Self-mixing ratio" denotes the percentage of images that are not mixing with other randomly paired images but mixup with an augmented version of themselves.}
\begin{tabular}{lcc}
\hline
\textbf{Self-mixing ratio} & \textbf{Clean Acc(\%)} & \textbf{Corruption Acc(\%)} \\
\hline
0\% & 82.86 & 56.15 \\
5\% & 82.83 & 58.83 \\
10\% & \textbf{83.50} & \textbf{59.02} \\
20\% & 83.02 & 58.97 \\
\hline
\end{tabular}
\label{table:self-mixing}
\end{table}

In our experiments, we also explore the concept of self-mixing, which refers to a particular case where an image does not undergo the usual mixup operation with another randomly paired image but instead blends with an augmented version of itself. This process can be controlled by the self-mixing ratio, denoting the percentage of images subject to self-mixing. 

Table~\ref{table:self-mixing} showcases the impact of the self-mixing ratio on the classification accuracy on both CIFAR-100 and CIFAR-100-C datasets when employing the ResNeXt-50 model. The results illustrate a notable trend: a 10\% self-mixing ratio leads to improvements in the classification performance, especially on the CIFAR-100-C dataset, which consists of corrupted versions of the original images. The improvement on CIFAR-100-C indicates that self-mixing contributes significantly to the model's robustness against various corruptions and perturbations. By incorporating self-mixing, our model gets exposed to a form of noise, thereby mimicking the potential real-world scenarios more effectively and enhancing the model's ability to generalize. The noise introduced via self-mixing could be viewed as another unique variant of the data augmentation, further justifying the importance of diverse augmentation strategies in improving the performance and robustness of the model.

\section{Conclusion}

In conclusion, our work in this paper has provided a significant contribution towards the development and understanding of Multi-layered Augmented Mixup (MiAMix). By reimagining the design of GMix, we have introduced an augmented form, AGMix, that leverages the Gaussian kernel's flexibility to produce a diversified range of mixing outputs. Additionally, we have devised an innovative method for sampling the mixing ratio when dealing with multiple mixing masks. Most crucially, we have proposed a novel approach for MSDA that incorporates various stages, namely: random sample pairing, mixing methods and ratios sampling, the generation and augmentation of mixing masks, and the output of mixed samples. By unifying these stages into a cohesive framework—MiAMix—we have constructed a search space replete with diverse hyper-parameters. This multi-stage approach offers a more diversified and dynamic way to apply data augmentation, potentially leading to improved model performance and better generalization on unseen data. Importantly, our methods do not incur excessive computational cost and can be seamlessly integrated into established training pipelines, making them practically viable. Furthermore, the versatile nature of MiAMix allows for future adaptations and improvements, promising an exciting path for the continuous evolution of data augmentation techniques. Given these advantages, we are optimistic about the potential of MiAMix to significantly influence and shape the field of machine learning, thereby enabling more robust and efficient model training processes.

\bibliographystyle{unsrt}
\bibliography{reference}
\end{document}